\documentclass[letterpaper, 10 pt, conference]{ieeeconf}
\IEEEoverridecommandlockouts    % This command is only needed if
\usepackage{graphics}           
\usepackage{times}              
\usepackage{amsmath}            
\usepackage{amssymb}            
\usepackage{graphicx}
\usepackage{algorithm}
\usepackage[noend]{algpseudocode}
\usepackage{booktabs}
\usepackage{color}
\definecolor{instructioncolor}{rgb}{.5,.5,.5}

\usepackage[font=small]{caption}

\def\secref#1{Sec.~\ref{#1}}
\def\figref#1{Fig.~\ref{#1}}
\def\tabref#1{Tab.~\ref{#1}}
\def\eqref#1{Eq.~(\ref{#1})}

\makeatletter
\usepackage{xspace}
\DeclareRobustCommand\onedot{\futurelet\@let@token\@onedot}
\def\@onedot{\ifx\@let@token.\else.\null\fi\xspace}
 
\def\ie{i.e\onedot}

\def\etal{{et al}\onedot}
\makeatother

\def\etalcite#1{\etal~\cite{#1}}

\usepackage{array}
\newcolumntype{L}[1]{>{\raggedright\let\newline\\\arraybackslash\hspace{0pt}}m{#1}}
\newcolumntype{C}[1]{>{\centering\let\newline\\\arraybackslash\hspace{0pt}}m{#1}}
\newcolumntype{R}[1]{>{\raggedleft\let\newline\\\arraybackslash\hspace{0pt}}m{#1}}

\usepackage{subfigure}

\title{\LARGE \bf Robust Onboard Localization in Changing Environments\\ Exploiting Text Spotting}

\author{Nicky Zimmerman \and Louis Wiesmann \and Tiziano Guadagnino \and Thomas L\"abe \and Jens Behley \and Cyrill Stachniss% <-this % stops a space
  \thanks{All authors are with the University of Bonn, Germany. Cyrill Stachniss is also with the Lamarr Institute for Machine Learning and Artificial Intelligence, Germany.}% 
  \thanks{This work has partially been funded by the Deutsche Forschungsgemeinschaft (DFG, German Research Foundation) under Germany's Excellence Strategy, EXC-2070 -- 390732324 -- PhenoRob and
  by the European Union's Horizon 2020 research and innovation programme under grant agreement No~101017008~(Harmony). $^1$ https://github.com/PRBonn/tmcl
  }%
}

\begin{document}
\maketitle
\thispagestyle{empty}
\pagestyle{empty}

%%%%%%%%%%%%%%%%%%%%%%%%%%%%%%%%%%%%%%%%%%%%%%%%%%%%%%%%%%%%%%%%%%%%%%%%%%%%%%%%
\begin{abstract}
  %
  %% WHY 
  % Use 1-2 not too long sentences, which clearly answer the WHY question: 
  % Why is this relevant, why should I care?
  Robust localization in a given map is a crucial component of most autonomous robots. 
  %% WHICH PROBLEM 
  % One sentence that explain the problem the paper addresses/investigates
  % Start with: In this paper, we address the problems of \dots
  In this paper, we address the problem of localizing in an indoor environment that changes and where prominent structures have no correspondence in the map built at a different point in time.  To overcome the discrepancy between the map and the observed environment caused by such changes, we exploit human-readable localization cues to assist localization. 
  These cues are readily available in most facilities and can be detected using RGB camera images by utilizing text spotting. We integrate these cues into a Monte Carlo localization framework using a particle filter that operates on 2D LiDAR scans and camera data.
  By this, we provide a robust localization solution for environments with structural changes and dynamics by humans walking.
  We evaluate our localization framework on multiple challenging indoor scenarios in an office environment. 
  The experiments suggest that our approach is robust to structural changes and can run on an onboard computer. 
  We release an open source implementation of our approach$^1$, which uses off-the-shelf text spotting, written in C++ with a ROS wrapper.

\end{abstract}

%%%%%%%%%%%%%%%%%%%%%%%%%%%%%%%%%%%%%%%%%%%%%%%%%%%%%%%%%%%%%%%%%%%%%%%%%%%%%%%%
\section{Introduction}
\label{sec:intro}

\begin{figure}[t]
  \centering  

  \includegraphics[width=0.85\columnwidth]{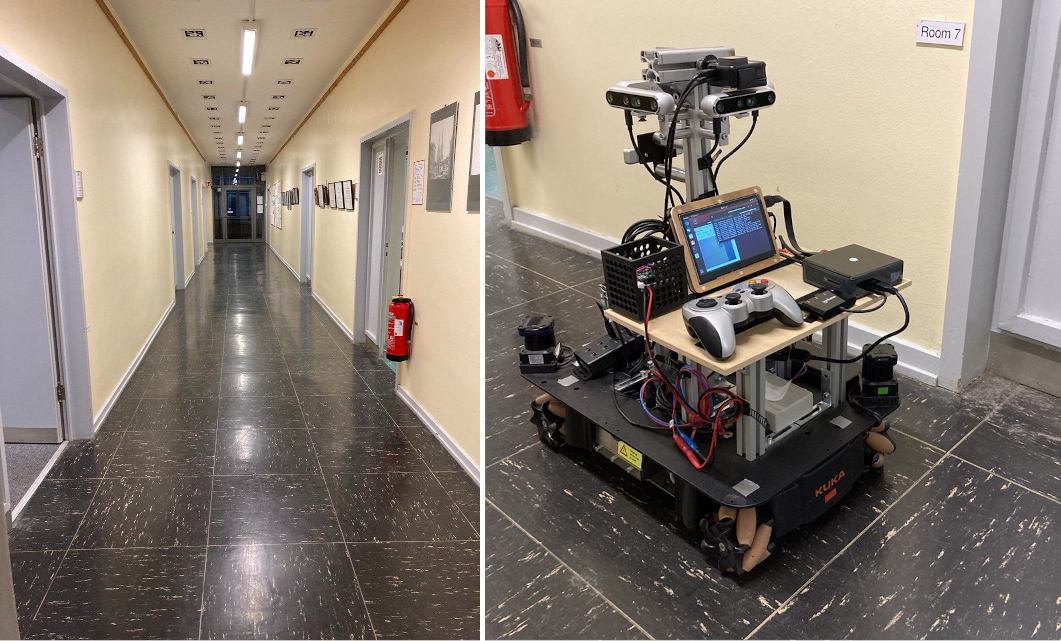}
  \includegraphics[width=0.92\columnwidth]{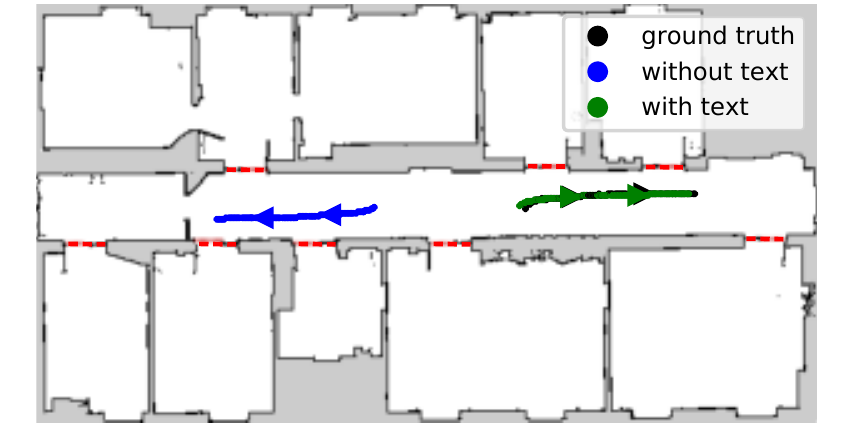}

  \caption{Top Left: The corridor in which the experiment took place in. Top right: The Kuka YouBot platform that was used for data collection, equipped with 2D LiDAR scanners and cameras that cover the complete $360^{\circ}$ field-of-view we utilize for text spotting. Bottom: The results of of localization in a corridor with closed doors (indicated by red lines), which are not reflected in the map, with and without textual cues.}
  \label{fig:motivation}
\end{figure}

%%%%%%%%%%%%%%%%%%% 
%% WHY: 
% First, answer the WHY question: Why is that relevant? Why should I be
% motivated to read the paper? Why should I care? (1 paragraph, 2-5 sentences)
%%%%%%%%%%%%%%%%%%%
%% WHICH PROBLEM
% Second, explain WHICH problem you are solving/address to solve.
Localization in a given map is a fundamental capability required by most autonomous robots operating in indoor environments, such as office or hospitals. 
These environments are often populated by people, also undergoing ``quasi-static'' changes such as closing of doors, objects temporarily standing at some place, or moved furniture that is not reflected in the given map that was recorded at a different point in time. Such changes, which we refer to as ``quasi-static'' in contrast to dynamic ones such as moving people, result in sensor observations that substantially differ from the map and can lead to localization failure, as illustrated in \figref{fig:motivation}, where closed doors in a corridor remove localization cues that can lead to ambiguities.

%%%%%%%%%%%%%%%%%%% 
%% HOW & WHAT
% Third, explain briefly how one can address the problem in general and mention 
% briefly what others/we before have done. Prepare the reader for your contribution 
% that comes in the next section (and not here!).
To overcome such localization challenges, readily available sources of information can be exploited to aid pose estimation. One example is using WiFi signal strength~\cite{ito2014icra} from existing access points to aid the  localization. Another example is using textual information that is part of the building infrastructure. Textual cues are often used by humans to navigate in the environment and are therefore available in most buildings designed for humans. With the recent advances in deep learning-based text recognition~\cite{shi2015arxiv}, we can reliably and efficiently decode textual content from images and utilize these hints in our localization approach. Surprisingly, there exist only a few approaches~\cite{cui2021iros}\cite{radwan2016icra} in the robotics community to exploit text spotting or optical character recognition~(OCR) for robot localization.
% Link to figure somewhere
% See \figref{fig:motivation} for an example.   
%%%%%%%%%%%%%%%%%%
%% MAIN CONTRIBUTION & WHAT FOLLOWS FROM THAT
% Explain your contribution in one paragraph. This is a very important paragraph. 
% Always start that paragraph with: ``The main contribution of this paper is''
The main contribution of this paper is a localization framework that integrates text spotting into a particle filter to improve localization. To this end, we build maps indicating the likelihood of detecting room numbers across the environment. The locations with high likelihood for successful detection are then used to inject particles when a known sign is detected. 
The textual cues allow us to globally localize with a small number of particles enabling online performance on mobile robots with limited computational resources. In our experiments, we show that our approach is able to
(i) localize in quasi-static environments,
(ii) localize in an environment with low dynamics,
(iii) localize in different maps types -- a featureless floor plan-like map, and LiDAR-based, feature rich map.
Furthermore, our approach runs online on an onboard computer.

%%%%%%%%%%%%%%%%%%%%%%%%%%%%%%%%%%%%%%%%%%%%%%%%%%%%%%%%%%%%%%%%%%%%%%%%%%%%%%%%
\section{Related Work}
\label{sec:related}

% Discuss the main related work and cite around 15-25 papers in sum. 
% The related work section should be approx. 1 column long, assuming 
% a 6-page paper.  Structure the section in paragraphs, grouping the 
% papers, and describing the key approaches with 1-2 sentences. If 
% applicable, describe the key difference to your approach at the end 
% of each paragraph briefly. Avoid adding subsections, al least for a 
% conference paper.     

% basic localization

Localization of mobile platforms is a well researched area in robotics~\cite{cadena2016tro}\cite{thrun2005probrobbook}\cite{zafari2019cst}. Probabilistic methods that estimate the robot's state have proven to be exceptionally robust, and include the extended Kalman filter (EKF) \cite{leonard1991tra}, Markov localization by Fox \etalcite{fox1999jair} and particle filters often referred to as Monte Carlo localization (MCL) by Dellaert \etalcite{dellaert1999icra}. These seminal works focused on localization using range sensors such as 2D LiDARs and sonars, as well as cameras. For cameras, the global localization task is framed under the visual place recognition framework, for which multiple algorithms have been proposed~\cite{bennewitz2006euros}\cite{cummins2008ijrr}.

% floor plans
Localization in feature-rich maps, often constructed by range sensors, is well-established \cite{moravec1985icra}. However, there are advantages for using sparse maps such as floor plans for localization. Floor plans are often available for buildings and do not require prior mapping with LiDARs or other range sensors. Their sparsity also means they do not need to be updated as frequently as detailed maps that include possibly moving objects, such as furniture. Their downside is their lack of details, which can render global localization challenging when faced with multiple identical rooms. Another issue is a possible discrepancy between the plans and the construction \cite{boniardi2017iros}. Boniardi \etalcite{boniardi2019iros} localize in floor plans with a camera by inferring the room layout and match it against the floor plans. Li \etalcite{li2020iros} introduce a new state variable, scale, to address the scale difference between floor plans and the actual structures. 

% fast dynamics
A problem arises when dynamic objects are detected in the scans and observations cannot be correctly matched to a given map. Sun \etalcite{sun2016iros} propose to detect those dynamic objects as outliers using a distance filter. Thrun \etalcite{thrun2005probrobbook} also incorporate the appearance of unexpected objects in the sensors model.
Another aspect of scene dynamics is changes that are longer-lasting and not as fast to appear and disappear like moving objects. These long-lasting changes can be closing and opening of passages, transferring large packages from one place to another and shifting of large furniture. Since those changes are more constant, standard filtering technique will fail to remove them. Stachniss and Burgard \cite{stachniss2005aaai} specifically address the case of closing and opening doors, by trying to detect areas of the map that can have different configurations and learn the possible environmental states in these corresponding areas. Another approach by Krajnik \etalcite{krajnik2016iros} try to capture periodic changes by representing every cell in the occupancy map as a periodic function. Another challenge are seasonal changes in an environment, which were addressed by Vysotska \etalcite{vysotska2016ral} and Milford \etalcite{milford2012icra}. In our approach, we do not assume to have prior knowledge on changes that may occur in the map, nor do we require long sequences of images to match against.

To tackle more general semi-permanent changes, Valencia \etalcite{valencia2014iros} suggest using multiple static maps, each corresponding to a different time scale. Biber \etalcite{biber2005rss} also propose to update a short-term map online. In the work of Tipaldi \etal~\cite{tipaldi2013ijrr}, a Hidden Markov model~\cite{baum1966ams} is assigned to every grid cell, creating dynamics occupancy grids that can be updated. These methods require continuous update of the map, while we handle changes without altering the map. 
% different modalities 

To assist global localization, additional modalities were considered. Ito \etalcite{ito2014icra} use WiFi signal strength to estimate the initial pose, based on signal strength maps that were previously constructed. Joho \etalcite{joho2009icra} suggest a sensor model for RFID that combines the likelihood of detecting a tag at a given pose and the likelihood of receiving a specific signal strength. We take inspiration from these papers for building our text likelihoods/priors but apply it for a different modality. %TODO, including fusion of many sensors \cite{wilbers2021phd}.

% text stuff 
Considerable amount of information is helping humans navigate, from publicly available maps to direction signs. Vysotska \etalcite{vysotska2016iros} use publicly available maps, like Open Street Map, to localize with LiDAR. However, exploitation of text for localization is not commonly explored. It was suggested by Radwan \etalcite{radwan2016icra} but considers outdoor environment and usage of Google Maps, while our approach tackles indoor environments.
Another implementation of text spotting in a MCL framework is presented by Cui \etalcite{cui2021iros}, who rely on text detection as its only sensor model. This differs from our work, which uses a 2D LiDAR-based sensor model and only leveraged text to improve global localization. The advantage of our method is that we are able to localize even in the absence of textual cues. Furthermore, in the work of Cui et al., text spotting is trained specifically for spotting parking space numbers, while we use a generic, off-the-shelf text spotting that performs well on a variety of textual cues~\cite{shi2015arxiv}.

 %% The approach by Stachniss \etalcite{stachniss2005aaai} aims at predicting \dots

%% BRIEFLY SUMMARIZE OWN CONTRIBUTION 

%%%%%%%%%%%%%%%%%%%%%%%%%%%%%%%%%%%%%%%%%%%%%%%%%%%%%%%%%%%%%%%%%%%%%%%%%%%%%%%%
\section{Our Approach}
\label{sec:approach}

Our goal is to globally localize in an indoor environment that can undergo significant structural changes using 2D LiDAR scanners, cameras and wheel odometry. In sum, we achieve this by building upon the Monte Carlo localization framework. To aid with global localization and recover from localization failures, we use a text spotting approach inferred from camera images to detect room numbers of an human oriented environment. 
To integrate the textual cues, we create text likelihood maps, which indicates the likelihood of detection of each room number as a function of the robot position. 
We inject particles corresponding to the locations suggested by the text likelihood.
%% Describe your approach. It is okay to divide the main section
%%  into a few subsections (e.g., 2-4 subsections).
\subsection{Monte Carlo Localization} 
\label{sec:MCL}

Monte Carlo localization \cite{dellaert1999icra} is a probabilistic method for estimating a robot's state $x_t$ given a map $m$ and sensor readings $z_t$ at time $t$. As we operate in an indoor environment, the robot's state $x_t$ is given by the 2D coordinates $(x,y)$ and the orientation $\theta \in [0,2\pi)$.  In our case an observation $z$ is composed of $K$ beams $z_k$ and the map $m$ is represented by an occupancy grid map~\cite{moravec1989sdsr}. 

We use a particle filter to represent the belief about the robot's state $p(x_t\mid z_{1:t}, m)$, where each particle \mbox{$s_t^{(i)}=\left(x_t^{(i)}, w_t^{(i)}\right)$} is represented by a state $x_t^{(i)}$ and a weight $w_t^{(i)}$. 
When odometry is available, successive states are sampled from a proposal distribution represented by holonomic motion model %(Forward-Sideways-Rotation) 
with odometry noise $\sigma_{\text{odom}} \in \mathbb{R}^3$. For each observation, each particle is weighted according to the likelihood of the observation given its state, \ie, $w_t^{(i)} = p(z_t\mid x_t^{(i)}, m)$.

As observation model $p(z_t\mid x_t,m)$, we use a beam-end model~\cite{thrun2005probrobbook}. The product of likelihood model assumes scan points are independent of each other. With the high angular resolution of our LiDAR this assumption does not hold. To address the overconfidence problem of the product of likelihood model, we decided to use the product of experts model \cite{miyagusuku2019ral}, where the weight of each particle is computed as the geometric mean of all scan points
\begin{align}
 p(z_t\mid x_t,m) &= \prod_{k=0}^K p(z_t^k\mid x_t,m)^{\frac{1}{K}},
\end{align}
where 
\begin{align}
 p(z_t^k\mid x_t,m) &=  \frac{1}{\sqrt{2 \pi \sigma_{\text{obs}}}} \exp{\left(-\frac{EDT(\hat{z}_t^k)^2}{2 \sigma^2}\right)}.
 \label{eq:beam_end_model}
\end{align} 

In \eqref{eq:beam_end_model}, $\hat{z}_t^k$ is the end point of the beam in the map~$m$, and EDT is the Euclidean distance transform~\cite{felzenszwalb2012toc} that indicates the distance to an occupied cell in the occupancy map. We truncated the EDT at a predefined maximal range,~$r_{\text{max}}$. 

For resampling, we chose low-variance resampling~\cite{thrun2005probrobbook} with an efficient sample size criteria~\cite{arulampalam2002tsp, bergman1999phd} of N/2, where N is the number of particles. Furthermore, our implementation of MCL \cite{dellaert1999icra} is asynchronous -- we sample from the motion model every time we get an odometry input, and we compute the weights whenever an observation is available and the robot traveled a predefined minimum distance $(d_{\text{xy}}, d_{\theta})$.

\begin{figure}[t]
  \centering
  \includegraphics[width=\columnwidth]{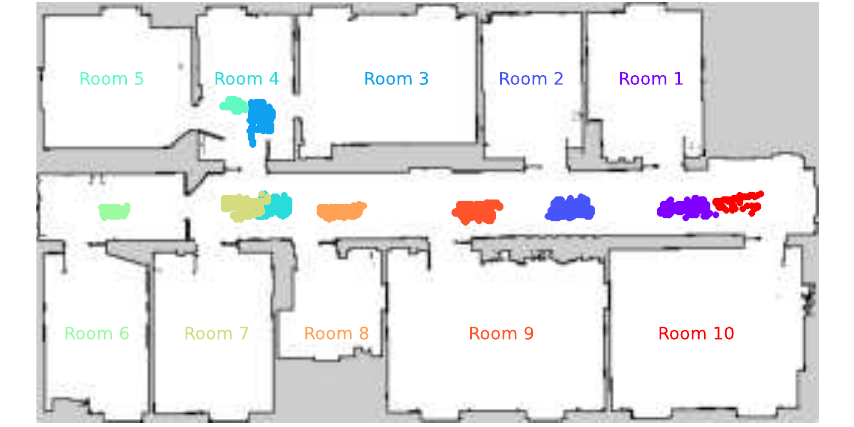}
  \caption{The text likelihood maps, based on the collected data, indicate the locations in which detection of each room number is likely. The likelihood maps are used for particle injection when a detection of a known text cues occurs.}

  \label{fig:text_maps}
\end{figure}

\subsection{Text Spotting} 
\label{sec:textspotting}

Text spotting can traditionally be split into text detection, \ie, localizing a bounding box that includes text, and text recognition, \ie, decoding the image patches extracted from the bounding boxes, to text. Text recognition is essentially a classification problem, therefore only the characters that are introduced during training can be inferred. The last decade's progress in object detection and text recognition allows us to use deep learning models for text spotting. 

\begin{figure}[t]
  \centering
  \subfigure[]{\includegraphics[width=0.49\columnwidth]{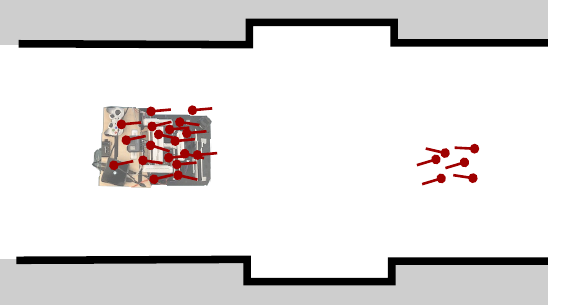}}
  \subfigure[]{\includegraphics[width=0.49\columnwidth]{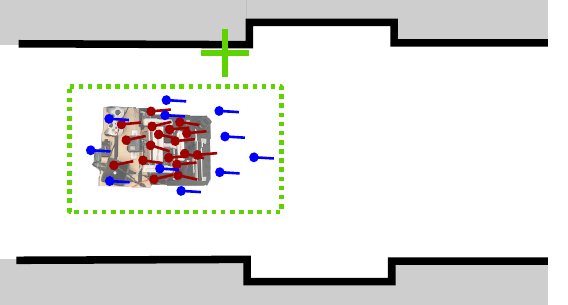}}

  \caption{Particle injection with text spotting. (a) Before detection, we have a situation with multi-modal distribution of particles (shown in red) as the corridor with closed doors is a symmetric situation that cannot be resolved just using the LiDAR scans. (b) With the first text detection (indicated by the green cross), we can inject new particles inside the bounding box extracted from the text map. We replace low weighted particles by new particles (shown in blue) that are uniformly distributed inside the corresponding bounding box of the text detection (shown by a dashed green line).}

  \label{fig:injection_after_text_detection}
\end{figure}

For the text spotting, we used the differentiable binarization text detector proposed by Liao \etalcite{liao2019arxiv}. The backbone is a ResNet18 \cite{he2016cvpr} neural network, which is powerful but also efficient enough to allow for fast inference. 

The text recognition model is based on the work of Shi \etalcite{shi2015arxiv}, who proposed the CRNN architecture, that combines convolution, recurrent and transcription layers. This model can handle text of arbitrary length, is end-to-end trainable without requiring fine-tuning and is relatively small while maintaining accuracy. We use four cameras, with a coverage of $360^{\circ}$, to spot text.
 
\subsection{Text Likelihood Maps}\label{sec:textmaps}
To incorporate text spotting into the MCL framework, we build a likelihood function of where the robot might detect a specific room number by collecting data that included image streams and the robot's pose. We apply the text spotting pipeline on the recorded images, assuming that the textual cues we are interested in follow a specific pattern (``Room X'') but it can generally be used for any textual content of interest. 

We compute 2D histograms for each room number, of locations where successful detections were made. The sampled locations give a sparse description of the text spotting likelihood, which we refer to as text likelihood maps (\figref{fig:text_maps}). As we are interested in a dense representation for the likelihood, we chose a simple strategy -- for each text tag, we compute an axis-aligned bounding box around all sampled locations where the detection rate is above threshold $\tau$ for this textual cue. We approximate the likelihood of text detection with an uniform distribution within the bounding box.

\subsection{Integration of Textual Cues} \label{sec:integration}
When a room number is detected, we store the room number and from which camera it was observed. Upon first detection, we inject particles into the corresponding area of the map (\figref{fig:injection_after_text_detection}). If the last detection was made from the same camera and of the same room number, we do not inject particles. 
The number of particles injected is defined by the injection ratio, $\rho$, the number of injected particles divided by the total number of particles.

In the injection process, for a particle filter with N particles, we first remove $\rho N$ particles with the lowest weights, and then inject an equal number of particles uniformly into the bounding box corresponding to that room number. The orientation $o_i$ of the injected particle $s_t^{(i)}$ depends on which camera spotted the text. We assumed that the camera detecting the text facing the room number at perpendicular angle. Thus, we inject particles with corresponding orientation and add Gaussian noise, $\sigma_{\text{inject}} = 0.05$. The injection ratio $\rho$ was chosen to be 0.5. A very high injection ratio could lead to localization failure if a wrong room number is detected. A low injection ratio has limited impact on the pose estimation. The injection of particles is done asynchronously, whenever a textual cue is available, and new particles are initialized with weight $w^{(i)} = \frac{1}{N}$.

\begin{figure}[t]
  \centering
  \includegraphics[width=0.7\columnwidth]{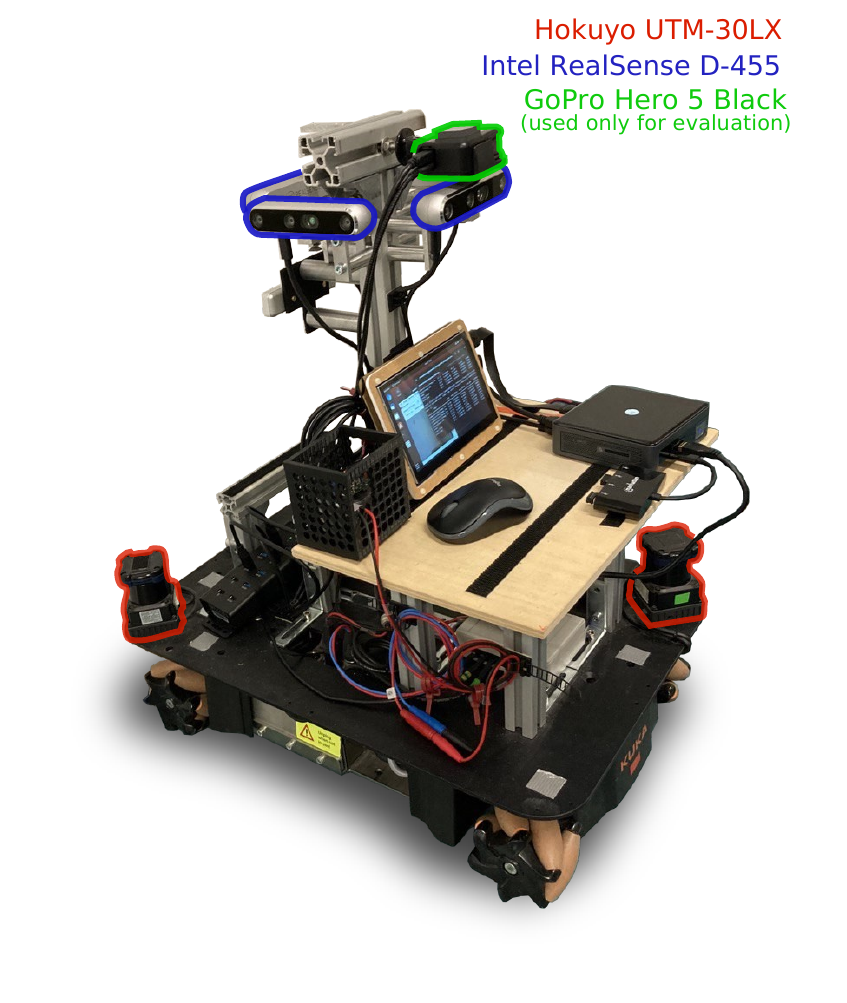}
  \caption{The data collection platform, an omnidirectional Kuka YouBot, with 2D LiDAR scanners (marked by a red outline) and with 4 cameras (marked by a blue outline) providing $360^{\circ}$ coverage. The up-ward facing camera (marked by a green outline) is only used for generating the ground truth via AprilTag detections.}
  \label{fig:youbot}
\end{figure}

%%%%%%%%%%%%%%%%%%%%%%%%%%%%%%%%%%%%%%%%%%%%%%%%%%%%%%%%%%%%%%%%%%%%%%%%%%%%%%%%
\section{Experimental Evaluation}
\label{sec:exp}

%% Repeat the main focus/objective with one single(!) sentence starting with:
%
The main focus of this work is an efficient, robust localization algorithm that leverage text information to better handle significant changes in the environment. We present our experiments to show the capabilities of our method and to support our claims, that we can 
(i) localize in quasi-static environments,
(ii) localize in an environment with low dynamics,
(iii) localize in different maps types -- a featureless floor plan-like map, and LiDAR-based, feature rich map. These capabilities can be run online on our robot.

\subsection{Experimental Setup}

%% If needed (and only then!) say also a few words about the experimental
%% setup, the datasets, and used parameters. You can use a separate subsection if you
%% want to put the focus on that but often that is not needed.}

To benchmark the performance of our approach, we recorded a dataset in an indoor office environment.  To this end, we equipped a Kuka YouBot platform with 2 Hokuyo UTM-30LX LiDAR sensors, 4 sideways-looking Intel RealSense RGB-D (D455), and an upward-looking GoPro Hero5 Black that is used only for evaluation purposes, as shown in \figref{fig:youbot}. We recorded the data including wheel odometry for different scenarios.

\begin{figure}[t]
  \centering 

  \subfigure[Sequences S1-S10]{\includegraphics[width=0.475\columnwidth]{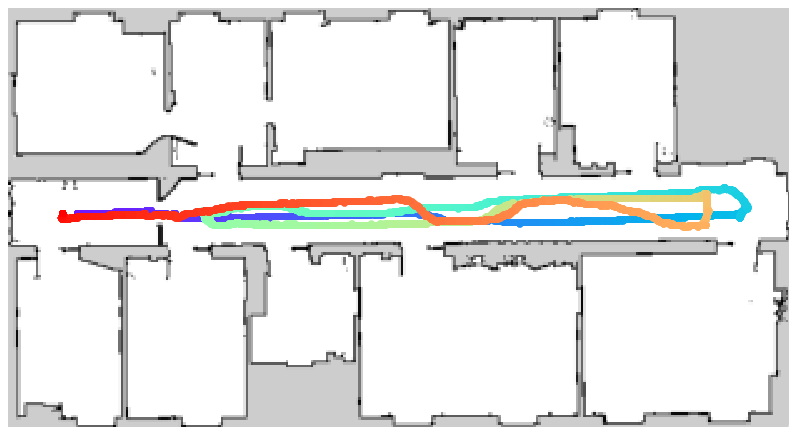}}
  \hfill
  \subfigure[Sequences D1.1-D1.5]{\includegraphics[width=0.475\columnwidth]{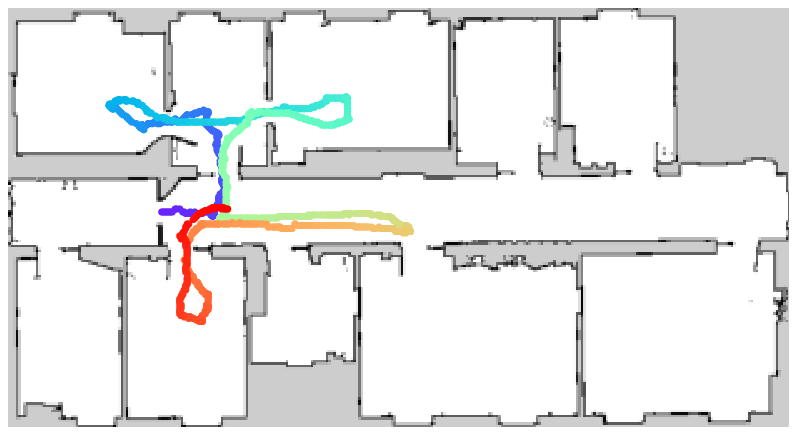}}

  \subfigure[Sequences D3.1-D3.5]{\includegraphics[width=0.475\columnwidth]{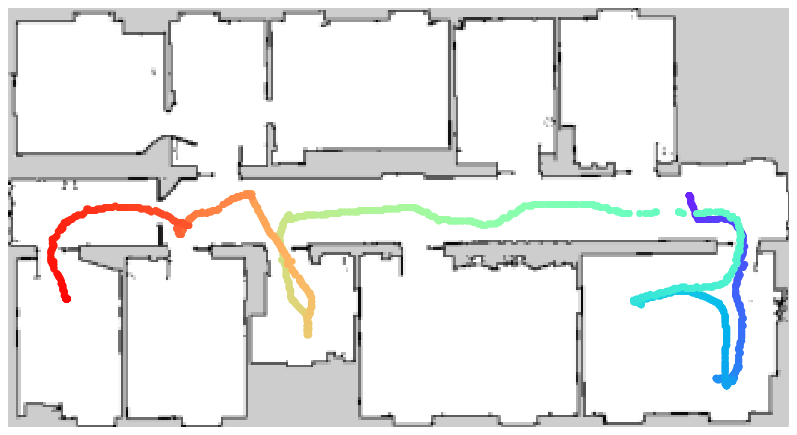}}
  \hfill
  \subfigure[Sequences D4.1-D4.4]{\includegraphics[width=0.475\columnwidth]{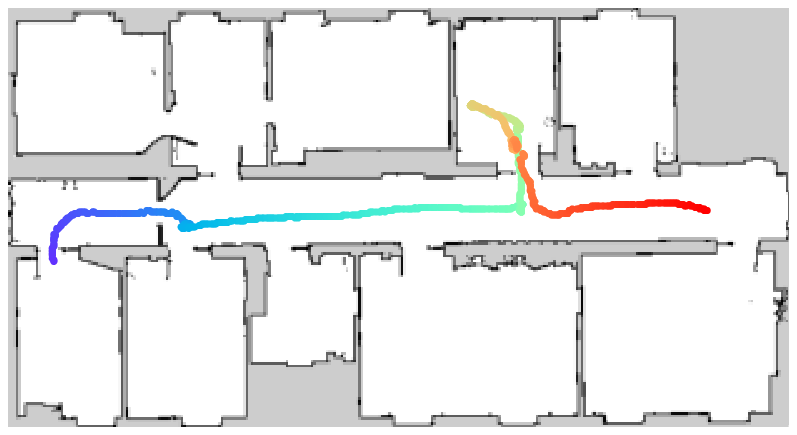}}

  \caption{Visualization of the different sequences used for evaluating our approach. Sequences S1-S10 correspond to the scenario where all doors are closed. Sequences D1-D4 were recorded with all the doors open, and with moderate amount of humans moving around. The color of the trajectory correspond to the time, where purple is the beginning and red corresponds to the end of the sequence.}
  \label{fig:gt_traj_plot}
\end{figure}

We recorded different scenarios. A long recording was made in the corridor with all doors closed, and sequences S1-S10 are randomly sampled from that data. Similarly, the sequences starting with D are sampled from recordings D1-D4, where doors were open but contain fast-moving dynamics. We include a plot of the trajectory of the scenarios in \figref{fig:gt_traj_plot}.

To determine the ground truth pose of the robot, we use precisely localized AprilTags \cite{olson2011icra-aara}, densely placed on the ceiling of every room and corridor. The AprilTags were detected using an up-facing camera that is used solely for this purpose. The AprilTags allow to continuously and accurately localize the robot with the dedicated sensor even under dynamic changes.

\begin{figure*}[t]
  \centering
  \subfigure[Map constructed by horizontally slicing a 3D point cloud captured with a FARO Focus X130 terrestrial laser scanner.]{\includegraphics[width=0.3\textwidth]{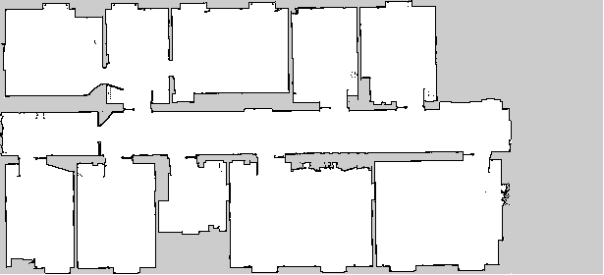}}
  \subfigure[Occupancy grid map from GMapping~\cite{grisetti2005icra} that was aligned to the FARO scan.]{\includegraphics[width=0.3\textwidth]{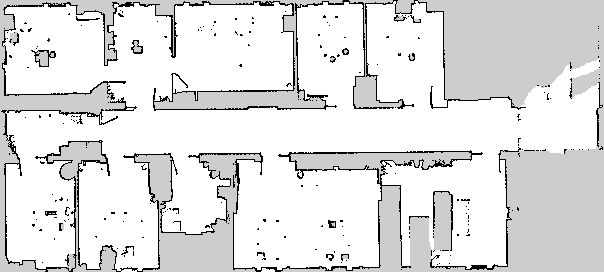}}
  \subfigure[Occupancy grid map from GMapping~\cite{grisetti2005icra} that was aligned to the FARO scan, based on the recordings from the corridor scenario.]{\includegraphics[width=0.3\textwidth]{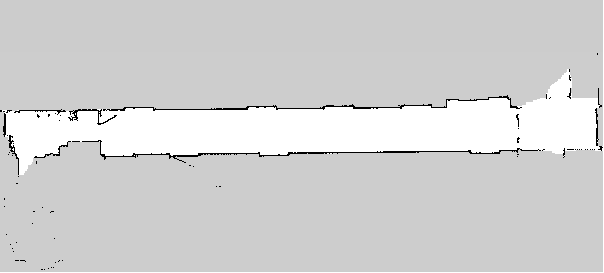}}

  \caption{Different maps used in the experiments: (a)~floor plan-like maps and (b)~LiDAR-based maps.~(c)~map built using GMapping, based on the recordings from the corridor scenario, which significantly deviates from the maps provided for localization.}
  \label{fig:two_maps} 
\end{figure*}

We explore two map representations, a floor plan-like map and a 2D LiDAR-based occupancy grid produced by GMapping~\cite{grisetti2005icra}, both illustrated in \figref{fig:two_maps}. The sparse, floor plan-like map was extracted from a high resolution terrestrial FARO laser scan, by slicing the dense point cloud at a fixed height. For all experiments, we use a map resolution of 0.05\,m/cell and the parameters specified in \tabref{tab:paramters}. 

\begin{table}[t]
  \caption{Algorithm parameters}
   \centering
   \setlength\tabcolsep{4.5pt}
 \begin{tabular}{ccccccc}\toprule
 $\sigma_{\text{odom}}$  &  $\sigma_{\text{obs}}$ & $r_{\text{max}}$ & $ \tau $ & $\rho$ & $d_{\text{xy}}$ & $d_{\theta}$\\ \midrule
 (0.02\,m, 0.02\,m, 0.02\,m) & 2.0 & 15.0\,m & 0 & 0.5 & 0.05\,m & 0.05\,rad\\ \bottomrule
 \end{tabular}
 \label{tab:paramters}
\end{table}

\begin{table*}[t] 
  \caption{ATE after convergence, for each sequence for the corridor scenario, using the \textbf{sparse map} with 300 particles. Angular error in radians / translational error in meters.}
  \centering 
  \resizebox{\textwidth}{!}{
    \setlength\tabcolsep{4.5pt}
\begin{tabular}{ccccccccccc}\toprule

Method        & S1 & S2 & S3 & S4 & S5 & S6 & S7 & S8 & S9 & S10\\ \midrule
AMCL &{-}/{-} &{-}/{-} &{-}/{-} &{-}/{-} &\textbf{0.01/0.110} &{-}/{-} &{-}/{-} &{-}/{-} &2.426/10.468 &{-}/{-} \\
MCL      &2.241/9.592 &{-}/{-} &0.287/0.594 &\textbf{0.045}/0.581 &{-}/{-} &1.795/7.803 &0.625/2.036 &2.465/11.550 &1.859/9.262 &1.011/1.640 \\
SM1     &0.405/2.329 &{-}/{-} &\textbf{0.01}/0.537 &\textbf{0.045}/0.563 &{-}/{-} &1.795/7.803 &0.625/2.022 &2.465/11.550 &1.855/9.393 &1.011/1.501 \\
SM2      &1.116/1.795 &0.777/2.597 &1.227/3.214 &0.706/2.611 &0.861/1.704 &1.600/3.782 &0.118/5.644 &0.321/0.563 &1.388/2.276 &1.148/1.466 \\
MCL+Text &\textbf{0.063/0.250} &\textbf{0.01/0.245} &0.063/\textbf{0.246} &0.063/\textbf{0.266} &\textbf{0.01}/0.246 &\textbf{0.179/0.369} &\textbf{0.045/0.221} &\textbf{0.077/0.343} &\textbf{0.493/0.332} &\textbf{0.01/0.184} \\
\bottomrule 
\end{tabular}
}
\label{tab:ate_s1s10_fmap} 
\end{table*}
\begin{table*}[t] 
  \caption{Errors averaged over the trajectory, after convergence, for the mostly static environment scenario, using the \textbf{sparse map} with 300 particles. Angular error in radians / translational error in meters.}
  \centering
  \resizebox{\textwidth}{!}{
\begin{tabular}{ccccccccccc}\toprule

Method        & D1.1 & D1.2 & D1.3 & D1.4 & D2.1 & D3.1 & D3.2 & D3.3 & D4.1 & D4.2\\ \midrule
AMCL   &2.022/6.031 &0.063/\textbf{0.135} &{-}/{-} &{-}/{-} &\textbf{0.010/0.095} &{-}/{-} &{-}/{-} &{-}/{-} &{-}/{-} &{-}/{-} \\
MCL   &{-}/{-} &0.413/0.690 &1.253/2.708 &0.893/1.961 &{-}/{-} &1.439/3.298 &2.284/4.743 &2.090/3.945 &{-}/{-} &0.703/1.021 \\
SM1   &{-}/{-} &1.970/3.637 &1.255/2.715 &0.893/1.961 &{-}/{-} &1.537/4.274 &2.284/4.743 &2.090/3.945 &{-}/{-} &1.007/6.035 \\
SM2   &1.315/3.942 &2.341/5.634 &1.346/4.275 &1.358/2.319 &{-}/{-} &{-}/{-} &1.628/3.336 &1.357/2.836 &1.524/5.708 &1.505/5.609 \\
MCL+Text   &\textbf{0.045/0.158} &\textbf{0.045}/0.175 &\textbf{0.077/0.182} &\textbf{0.010/0.152} &0.045/0.279 &\textbf{0.010/0.133} &\textbf{0.333/0.697} &\textbf{0.010/0.141} &\textbf{0.045/0.161} &\textbf{0.063/0.197} \\\bottomrule 
\end{tabular}
}
\label{tab:ate_d1d4_fmap} 
\end{table*}

\begin{table}[t] 
  \caption{Convergence time in seconds, for the corridor scenario, using the \textbf{sparse map}, with 300 particles. In parentheses, the length of the sequences in seconds.}
   \centering
   \resizebox{\columnwidth}{!}{
     \setlength\tabcolsep{1.5pt}
 \begin{tabular}{ccccccccccc}\toprule
 
 Method        & S1 & S2 & S3 & S4 & S5 & S6 & S7 & S8 & S9 & S10\\ 
               &\scriptsize{(234.4)}  & \scriptsize{(229.6)}  & \scriptsize{(220.8)}  & \scriptsize{(212.1)}  & \scriptsize{(203.2)}  & \scriptsize{(187.7)}  & \scriptsize{(176.3)}  & \scriptsize{(152.1)}  & \scriptsize{(145.4)}  & \scriptsize{(123.0)} \\\midrule
 AMCL &- &- &- &- &54.5 &- &- &- &19.6 &-\\
MCL       &99.3 &- &18.6 &\textbf{0.0}  &- &75.9 &148.8  &121.5  &\textbf{11.5} &3.2\\
SM1       &15.7 &221.9  &35.2 &\textbf{0.0}  &- &75.9 &148.8  &121.5  &\textbf{11.5} &3.2\\
SM2       &\textbf{0.0}  &\textbf{1.5}  &\textbf{4.7}  &\textbf{0.0}  &126.5  &56.5 &135.4  &\textbf{5.6}  &\textbf{11.5} &3.3\\
MCL+Text  &\textbf{0.0}  &1.7  &12.6 &\textbf{0.0}  &\textbf{10.9} &\textbf{11.2} &\textbf{50.2} &5.7  &\textbf{11.5} &\textbf{2.4}\\
 \bottomrule 
 \end{tabular}
 }
 \label{tab:conv_s1s10_fmap} 
 \end{table}   

As baseline, we compare against AMCL~\cite{pfaff2006springerstaradvanced}, which is a publicly available and highly-used ROS package for MCL-based localization, and our implementation of MCL that does not rely on textual cues. Additionally, we implemented two sensor models for integrating textual information into the MCL framework, referred to as SM1 and SM2. SM1 assigns all particles within the bounding box a high weight, \mbox{$w^k_t=1.0$}, and a low weight, \mbox{$w^k_t=0.1$}, to particles elsewhere. SM2 converts the bounding box into a likelihood map and the weight for each particle is proportional to a Gaussian applied on its distance from the bounding box, similar to \eqref{eq:beam_end_model}. All experiments were executed with 300 particles unless mentioned otherwise, and N particles are initialized uniformly across the map.

We consider two metrics, time to convergence and absolute trajectory error~(ATE) after convergence. We define convergence as the point where the prediction is within a distance of 0.5\,m from the ground truth pose. If convergence did not occur within the first $95\%$ of the sequence, then we consider it a failure, which is marked as ${-}/{-}$.

%% Note 1: It MUST be always crystal clear (a) WHY an experiment is there
%% (e.g., to support a claim, to show that the approaches useful for real-word
%% systems, to show the performance, or to provide a baseline comparison), (b)
%% WHAT it wants to show (which claim/property exactly), and (c) HOW it aims at 
%% showing this. This is ESSENTIAL for a good evaluation. Think about when BEFORE
%% designing an experiment.  IMPORTANT: Every experiment MUST start with something 
%% like:  The next experiment is presented to show \dots and thus for supporting our 
%% first claim.

\subsection{Localization under Changes using a Sparse Map}
\begin{figure}[t] 
  \centering
  \includegraphics[width=0.87\columnwidth]{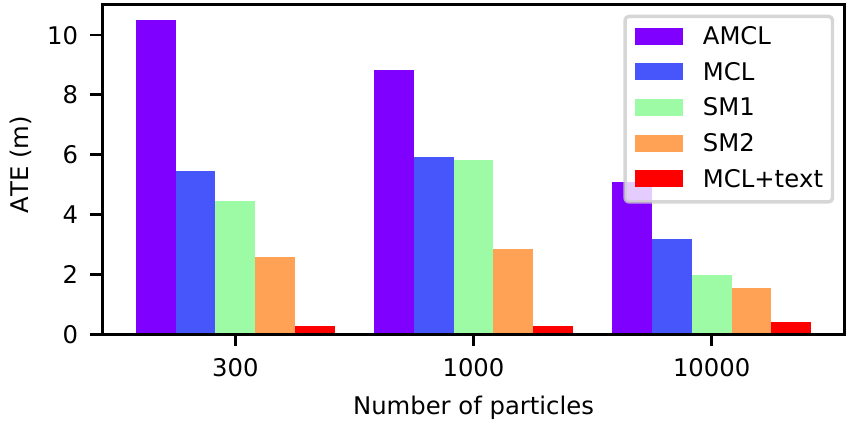}
  \caption{ATE~(xy) averaged over sequences S1-S10 as a function of the number of particles used in the particle filter, for the different methods method. The error for MCL+Text is similar across large range of particle set sizes, exhibiting the robustness of our approach.}
  \label{fig:ParticleNumAVG}
\end{figure}

%% First experiment - most impressive, important or the most important
%% claim supporting experiments comes first.
The first experiment evaluates the performance of our approach and supports the claim that we can localize in changing environment using floor plan-like maps. It is conducted on sequences recorded in a long corridor with all doors closed, while in the map these doors are all open, and it supports our claim of robust localization in face of quasi-static changes. We consider 10 sequences (S1-S10), each sequence starting at the a different location along the corridor. We evaluate the time to convergence and ATE for this challenging scenario on the 10 sequences. As can be seen in \tabref{tab:ate_s1s10_fmap} and \tabref{tab:conv_s1s10_fmap}, our text-enriched method converges quickly, and outperformed the baselines in all sequences. When the map not longer reflect the environment, it is expected that classic MCL implementations would perform poorly. For text spotting sensor model to affect the pose estimation, a particle must be in the close vicinity of a specific text likelihood bounding box. With relative low number of particles, such as 300, it is unlikely to have enough particles in such a small area. Therefore, the sensor model methods have limited contribution to global localization compared to particle injection. 
The MCL+Text method also shows exceptional robustness when reducing the number of particles in the filter, as can be seen in \figref{fig:ParticleNumAVG}. The ATE for our approach is slightly larger for 10,000 particles, due to the formation of multi-modal hypotheses caused by the symmetry of the corridor.

\begin{table}[t] 
  \caption{Convergence time in seconds, for the mostly static environment scenario, using the \textbf{sparse map}, with 300 particles. In parentheses, the length of the sequences in seconds.}
   \centering
   \resizebox{\columnwidth}{!}{
    \setlength\tabcolsep{2.5pt}
 \begin{tabular}{ccccccccccc}\toprule
 
 Method        & D1.1 & D1.2 & D1.3 & D1.4 & D2.1 & D3.1 & D3.2 & D3.3 & D4.1 & D4.2\\ 
 &\scriptsize{(171.4)}  &\scriptsize{(162.4)}  &\scriptsize{(144.8)}  &\scriptsize{(130.5)}  &\scriptsize{(78.2)} &\scriptsize{(177.7)}  &\scriptsize{(160.6)}  &\scriptsize{(147.7)}  &\scriptsize{(120.0)}  &\scriptsize{(100.4)}\\\midrule
 AMCL &112.4  &8.7  &- &- &10.6 &- &- &- &- &-\\
MCL       &167.6  &9.6  &80.9 &53.6 &- &69.2 &\textbf{55.2} &17.7 &- &7.4\\
SM1       &167.6  &136.7  &80.9 &53.6 &- &70.6 &\textbf{55.2} &17.7 &- &1.8\\
SM2       &\textbf{2.1}  &106.4  &\textbf{0.2}  &18.4 &- &- &\textbf{55.2} &22.9 &\textbf{20.4} &\textbf{0.0}\\
MCL+Text  &2.2  &\textbf{2.0}  &0.7  &\textbf{16.3} &\textbf{4.8}  &\textbf{64.6} &55.9 &\textbf{14.9} &36.4 &\textbf{0.0}\\\bottomrule 
 \end{tabular}
 }
 \label{tab:conv_d1d4_fmap} 
 \end{table}

\begin{table*}[t] 
  \caption{Errors averaged over the trajectory, after convergence, for each sequence for the corridor scenario, using the \textbf{GMapping map} with 300 particles. Angular error in radians / translational error in meters.}
  \centering
  \resizebox{\textwidth}{!}{
\begin{tabular}{ccccccccccc}\toprule

Method        & S1 & S2 & S3 & S4 & S5 & S6 & S7 & S8 & S9 & S10\\ \midrule
AMCL   &{-}/{-} &{-}/{-} &\textbf{0.010/0.087} &{-}/{-} &{-}/{-} &{-}/{-} &{-}/{-} &2.436/9.151 &2.420/10.756 &{-}/{-} \\
MCL   &{-}/{-} &2.399/9.822 &0.941/6.763 &1.352/7.113 &\textbf{0.010}/0.242 &\textbf{0.010}/0.547 &2.176/7.869 &{-}/{-} &2.600/9.385 &1.280/5.901 \\
SM1   &\textbf{0.010}/0.294 &0.601/3.727 &0.941/6.781 &1.407/7.486 &2.480/11.063 &\textbf{0.010}/0.525 &2.176/7.869 &{-}/{-} &2.600/9.385 &1.918/7.056 \\
SM2   &1.347/4.181 &1.344/4.644 &2.014/5.334 &1.618/4.257 &1.002/3.598 &1.566/4.100 &\textbf{0.499}/1.915 &0.476/1.936 &2.312/8.422 &1.414/4.625 \\
MCL+Text   &0.063/\textbf{0.191} &\textbf{0.063/0.203} &0.063/0.216 &\textbf{0.063/0.228} &0.063/\textbf{0.171} &0.045/\textbf{0.293} &0.632/\textbf{1.788} &\textbf{0.010/0.192} &\textbf{0.697/0.920} &\textbf{0.205/0.196} \\
\bottomrule 
\end{tabular}
}
\label{tab:ate_s1s10_gmap} 
\end{table*}

\begin{table*}[t] 
  \caption{Errors averaged over the trajectory, after convergence, for mostly static scenarios, using the \textbf{GMapping map} with 300 particles. Angular error in radians / translational error in meters.}
  \centering
  \resizebox{\textwidth}{!}{
\begin{tabular}{ccccccccccc}\toprule

Method        & D1.1 & D1.2 & D1.3 & D1.4 & D2.1 & D3.1 & D3.2 & D3.3 & D4.1 & D4.2\\ \midrule
AMCL   &0.547/3.490 &{-}/{-} &{-}/{-} &{-}/{-} &{-}/{-} &2.389/17.461 &{-}/{-} &{-}/{-} &{-}/{-} &{-}/{-} \\
MCL   &1.714/2.794 &1.895/3.982 &0.215/1.230 &0.495/1.144 &0.262/1.864 &1.424/3.427 &0.885/10.403 &0.812/3.825 &0.991/1.190 &0.632/9.345 \\
SM1   &0.425/3.683 &1.895/3.982 &0.215/1.230 &0.495/1.145 &0.265/1.828 &1.422/3.453 &0.045/0.225 &1.599/5.664 &0.991/1.190 &1.335/7.131 \\
SM2   &0.778/1.843 &1.881/4.875 &1.169/2.799 &1.083/1.772 &0.118/0.347 &1.497/5.087 &1.593/2.155 &1.628/5.366 &1.404/6.117 &0.704/1.477 \\
MCL+Text   &\textbf{0.010/0.109} &\textbf{0.010/0.109} &\textbf{0.077/0.099} &\textbf{0.010/0.116} &\textbf{0.045/0.161} &\textbf{0.010/0.156} &\textbf{0.010/0.169} &\textbf{0.045/0.165} &\textbf{0.063/0.160} &\textbf{0.045/0.131} \\
\bottomrule 
\end{tabular}
}
\label{tab:ate_d1d4_gmap} 
\end{table*}

\begin{table}[t] 
  \caption{Convergence time in seconds, for the corridor scenario, using the \textbf{GMapping map}, with 300 particles. In parentheses, the length of the sequences in seconds.}
   \centering
   \resizebox{\columnwidth}{!}{
    \setlength\tabcolsep{2.5pt}
 \begin{tabular}{ccccccccccc}\toprule
 
 Method        & S1 & S2 & S3 & S4 & S5 & S6 & S7 & S8 & S9 & S10\\ 
 &\scriptsize{(234.4)}  &\scriptsize{(229.6)}  &\scriptsize{(220.8)}  &\scriptsize{(212.1)}  &\scriptsize{(203.2)}  &\scriptsize{(187.7)}  &\scriptsize{(176.3)}  &\scriptsize{(152.1)}  &\scriptsize{(145.4)}  &\scriptsize{(123.0)}\\\midrule
AMCL &- &- &113.0  &- &- &- &- &\textbf{22.8} &19.7 &-\\
MCL       &- &29.5 &21.8 &\textbf{0.2}  &159.1  &122.8  &\textbf{7.9}  &- &\textbf{8.9}  &\textbf{1.1}\\
SM1       &18.1 &9.7  &21.8 &\textbf{0.2}  &173.2  &122.2  &\textbf{7.9}  &- &\textbf{8.9}  &\textbf{1.1}\\
SM2       &6.5  &1.7  &\textbf{3.2}  &\textbf{0.2}  &114.1  &10.9 &\textbf{7.9}  &47.8 &\textbf{8.9}  &\textbf{1.1}\\
MCL+Text  &\textbf{0.0}  &\textbf{1.5}  &3.6  &\textbf{0.2}  &\textbf{8.4}  &\textbf{9.7}  &\textbf{7.9}  &31.2 &\textbf{8.9}  &\textbf{1.1}\\
\bottomrule 
 \end{tabular}
 }
 \label{tab:conv_s1s10_gmap} 
 \end{table}

\subsection{Localization under Few Dynamics using a Sparse Map}
%% Second experiment - could be a comparison to a baseline method,
%% quality analysis or similar

The second experiment is presented to support the claim that our approach is able to localize in a floor plan-like map (not built using the robot's sensors) when the environment is mostly static. Recordings D1-D4 are taken across the lab, through different office rooms, with a small number of people moving around. In all sequences, all doors are open, and the environment is similar to the map. 

This experiment considers localization in a feature-sparse map and in the presence of low dynamics. This presents its own challenges even in a mostly unchanging environment. As seen in \tabref{tab:ate_d1d4_fmap} and \tabref{tab:conv_d1d4_fmap} our text-enriched method performs best. Despite having the doors open, these scenarios include movement in a corridor with very high symmetry. Textual cues can contribute to breaking such symmetries. In addition, there are many details such as furniture, that are not part of the sparse map and can affect the accuracy of LiDAR-only localization. While SM2 shows a rather promising convergence time, the impact of the text-based sensor model is milder than particle injection, leading to divergence later on, and a large ATE.

\subsection{Localization using LiDAR-Based Map Built with the Robot's Sensors}

The third experiment is presented to support the claim that our approach is able to localize in a LiDAR-based map, when the environment is structurally changing or when there are a few dynamics in the scene.
To ensure our algorithm works sufficiently well in LiDAR-based maps, we constructed a GMapping map based on 2D LiDAR scans. While this map is more detailed, the recordings were made across several weeks, resulting in some differences between the map and the environment. It is still difficult to localize globally with only 300 particles in a big scene, therefore our text-guided method enjoys an advantage. Our approach outperformed the baselines also in corridor scenario, as can be seen in \tabref{tab:ate_s1s10_gmap}. While the sensor model methods manage to converge in a timely manner (\tabref{tab:conv_s1s10_gmap}), they are less stable than our injection technique and result in greater ATE. Similarly, for the mostly static scenario, our approach achieves the best ATE overall (\tabref{tab:ate_d1d4_gmap}), in addition to its fast convergence, displayed in \tabref{tab:conv_d1d4_gmap}.

  \begin{table}[t] 
 \caption{Convergence time in seconds, for the mostly static scenarios, using the \textbf{GMapping map}, with 300 particles. In parentheses, the length of the sequences in seconds.}
  \centering
  \resizebox{\columnwidth}{!}{
     \setlength\tabcolsep{2.5pt}
\begin{tabular}{ccccccccccc}\toprule

Method        & D1.1 & D1.2 & D1.3 & D1.4 & D2.1 & D3.1 & D3.2 & D3.3 & D4.1 & D4.2\\ 
&\scriptsize{(171.4)}  &\scriptsize{(162.4)}  &\scriptsize{(144.8)}  &\scriptsize{(130.5)}  &\scriptsize{(78.2)} &\scriptsize{(177.7)}  &\scriptsize{(160.6)}  &\scriptsize{(147.7)}  &\scriptsize{(120.0)}  &\scriptsize{(100.4)}\\\midrule
AMCL  &\textbf{0.0}  &- &- &- &- &\textbf{35.2} &- &- &- &-\\
MCL       &111.2  &10.1 &30.7 &50.9 &21.5 &67.8 &133.4  &31.0 &51.6 &64.2\\
SM1       &115.7  &10.1 &30.7 &50.9 &21.5 &67.8 &87.2 &39.2 &51.6 &8.3\\
SM2       &2.2  &73.9 &30.2 &16.2 &7.1  &72.0 &\textbf{50.3} &\textbf{15.6} &26.5 &0.5\\
MCL+Text  &2.1  &\textbf{1.8}  &\textbf{0.2}  &\textbf{16.0} &\textbf{4.7}  &71.5 &50.6 &19.3 &\textbf{21.4} &\textbf{0.0}\\
\bottomrule 
\end{tabular}
}
\label{tab:conv_d1d4_gmap} 
\end{table}

%%%%%%%%%%%%%%%%%%%%%%%%
\subsection{Runtime}

%% Runtime experiment - it is often one of the last experiments unless
%% online processing/speed is the key contribution
The next set of experiments has been conducted to support our fourth claim that our
approach runs fast enough to execute online on the robot in real-time. We, therefore, tested our approach once using a Dell Precision-3640-Tower and once on an Intel NUC10i7FNK, which we have on our YouBot. The Dell PC has 20 CPU cores at 3.70\,GHz and 64\,GB of RAM. The Intel NUC has 12 CPU cores at 1.10\,GHz and 16\,GB of RAM.

Text spotting on the NUC runs at an average of 167\,ms, and on the desktop 100\,ms. \tabref{tab:speed} summarizes the runtime results for our approach. The
numbers support our fourth claim, namely that the computations can be executed fast and in an online fashion. 

\begin{table}[t] 
 \caption{Average inference time in ms for the sensor model on the NUC as a function of the number of particles.}
  \centering
  \resizebox{\columnwidth}{!}{
\begin{tabular}{ccccc}\toprule
        & 300 & 500 & 1\,000 & 10\,000 \\ \midrule
NUC10i7FNK    &  30  &   51 & 106 & 1027    \\
Dell Precision-3640-Tower & 24 & 40 & 80 & 793\\ \bottomrule
\end{tabular}
}
 \label{tab:speed}
\end{table}

\subsection{Ablation Study}
\begin{figure}[t]
  \centering
  \subfigure[ATE~(xy) averaged over sequences S1-S10.]{\includegraphics[width=0.875\columnwidth]{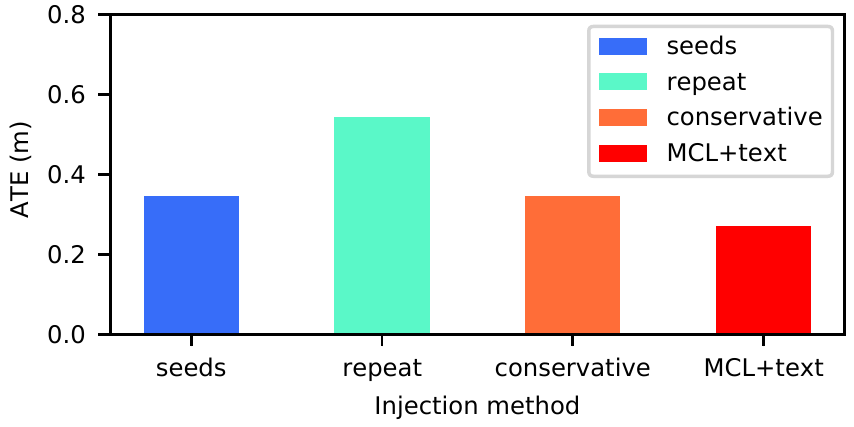}}
  \subfigure[Convergence time for sequences S1-S10.]{\includegraphics[width=0.875\columnwidth]{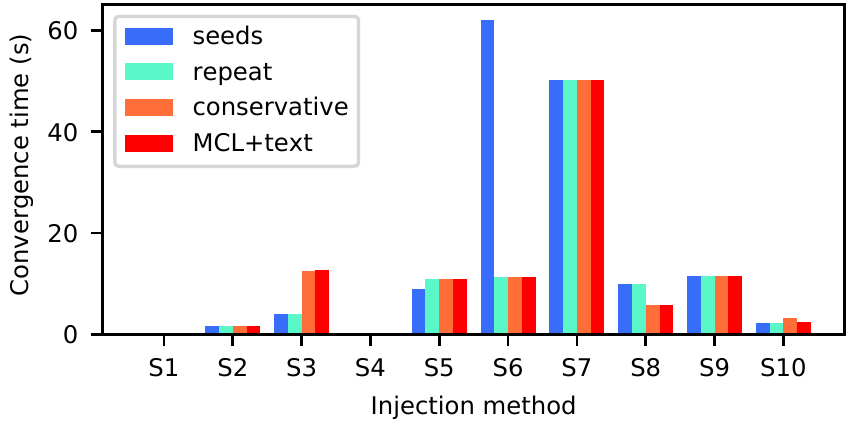}}

  \caption{Results for the ablation study exploring different injection strategies, with the sparse map and 300 particles.}
  \label{fig:ablation}
\end{figure}

Additionally, we conducted an ablation study to identify the best way of integrating the textual hints into our MCL framework. In addition to our MCL+Text method, we also explored the following strategies for injecting particles:
\begin{enumerate}
\item Seed locations: Specific hand-picked locations in the map, which correspond to room number locations, and are used to sample particles around them with a predefined covariance.
\item Repeat: Using the text likelihood maps, described in \secref{sec:textmaps}, we compute a bounding box for each room number plate, and inject particles in that area for every room number detection. If we have multiple consecutive detections of a room number from the same camera, we inject particles each time. 
\item Conservative: Using the text likelihood maps, we compute a bounding box for each room number plate, and inject particles in that area only once, if the filter's pose estimation mean does not lie in the bounding box. If the mean pose of MCL is within the bounding box, the filter is in line with the tag observations, and we do not inject particles. If we have multiple consecutive detections of a room number from the same camera, we inject particles only in the first detection. 
\end{enumerate}
 As can be seen in \figref{fig:ablation}, MCL+Text outperforms the other text-guided methods. MCL+Text also converges faster than the other text-guided methods. 
 % As can be seen in \figref{fig:ablation_ATE}, MCL+Text outperforms the other text-guided methods. Further analysis shows (\figref{fig:ablation_conv}) that MCL+Text also converges faster than the other text-guided methods. 

% \begin{figure}[t] 
%   \centering
%   \includegraphics[width=0.87\columnwidth]{pics/AblationAVG.pdf}
%   \caption{ATE~(xy) averaged over sequences S1-S10 for different injection strategies, with the sparse map and 300 particles.}
%   \label{fig:ablation_ATE}
% \end{figure}

% \begin{figure}[t]
%   \centering
%   \includegraphics[width=0.87\columnwidth]{pics/AblationConv.pdf}
%   \caption{Convergence time for sequences S1-S10 for different injection strategies, with the sparse map and 300 particles.}
%   \label{fig:ablation_conv}
% \end{figure}

%%%%%%%%%%%%%%%%%%%%%%%%%%%%%%%%%%%%%%%%%%%%%%%%%%%%%%%%%%%%%%%%%%%%%%%%%%%%%%%%
\section{Conclusion}
\label{sec:conclusion}

In this paper, we presented a novel approach to localize a robot in environments that deviate significantly from the provided map, as illustrated in \figref{fig:two_maps}, due to changes in the scene.
Our method exploits the readily available human-readable textual cues that assist humans in navigation. 
This allows us to successfully overcome localization failure in the cases where critical changes to the layout differ greatly from the map. 
We implemented and evaluated our approach on a dataset collected strictly for simulating such structural alterations, and provided comparisons to other existing techniques and supported
all claims made in this paper. The experiments suggest that incorporating human-readable localization cues in mobile robot localization systems provides considerable improvement in robustness. 

%%%%%%%%%%%%%%%%%%%%%%%%%%%%%%%%%%%%%%%%%%%%%%%%%%%%%%%%%%%%%%%%%%%%%%%%%%%%%%%%
%% Future work: Use only if applicable -- but if so, use the following
%% sentence to start:
% Despite these encouraging results, there is further space for improvements. 

%%%%%%%%%%%%%%%%%%%%%%%%%%%%%%%%%%%%%%%%%%%%%%%%%%%%%%%%%%%%%%%%%%%%%%%%%%%%%%%%
% Only if applicable
\section*{Acknowledgments}
We thank Holger Milz, Michael Plech, and Ralf Becker for their contribution in assembling our mobile platform. 
\bibliographystyle{plain_abbrv}

% All new citations should go to new.bib. The file glorified.bib should go
% be the one from the ipb server. After paper or related work has been
% written merge the entries from new.bib to glorified.bib ON THE SERVER,
% replace the glorified.bib in this repository and empty the new.bib
\bibliography{glorified,new}

\end{document}